# Hierarchical generative modelling for autonomous robots

Kai Yuan[1,4], Noor Sajid[2,4], Karl Friston[2] and Zhibin Li[3*]

[1] Embodied AI Lab, Intel Labs, Germany
[2] Wellcome Centre for Human Neuroimaging, Queen Square Institute of Neurology, University College London, UK
[3] Department of Computer Science and UCL Robotics Institute, University College London, UK
[4] These authors contributed equally.

* Corresponding author: alex.li@ucl.ac.uk.

**ABSTRACT**

Humans can produce complex whole-body motions when interacting with their surroundings, by planning, executing and combining individual limb movements. We investigated this fundamental aspect of motor control in the setting of autonomous robotic operations. We approach this problem by hierarchical generative modelling equipped with multi-level planning—for autonomous task completion—that mimics the deep temporal architecture of human motor control. Here, temporal depth refers to the nested time scales at which successive levels of a forward or generative model unfold, for example, delivering an object requires a global plan to contextualise the fast coordination of multiple local movements of limbs. This separation of temporal scales also motivates robotics and control. Specifically, to achieve versatile sensorimotor control, it is advantageous to hierarchically structure the planning and low-level motor control of individual limbs. We use numerical and physical simulation to conduct experiments and to establish the efficacy of this formulation. Using a hierarchical generative model, we show how a humanoid robot can autonomously complete a complex task that necessitates a holistic use of locomotion, manipulation, and grasping. Specifically, we demonstrate the ability of a humanoid robot that can retrieve and transport a box, open and walk through a door to reach the destination, approach and kick a football, while showing robust performance in presence of body damage and ground irregularities. Our findings demonstrated the effectiveness of using human-inspired motor control algorithms, and our method provides a viable hierarchical architecture for the autonomous completion of challenging goal-directed tasks.

# INTRODUCTION

Humans can control their bodies to produce intricate motor behaviours that align with their objectives, e.g., navigating in an environment with a mixed sequential use of legs and hands in a coherent manner. These tasks require the coordination of multiple processes, including motor planning and execution[1, 2,3,4]. To realise this coordination, human motor control unfolds at nested time scales at different levels of the neuronal hierarchy[5, 6], e.g., a high-level plan to arrive at a particular place can entail multiple, individual, reflexive low-level limb movements for walking. In the areas of robotics, hierarchical elements have been applied to control systems to achieve diverse motor behaviours[7,8]. The core principles to achieve human-like motor control have been derived and summarised previously[9], by relating these elements to the human nervous system.

In robotics, past research has been conducted to achieve similar capabilities as humans, such as assembly in aircraft manufacturing,[10] space missions[11], as well as the computational model of active inference for robust robot behaviours[12]. These have been achieved by using mainly three approaches: human commands, planning, and learning.

While human commands have been used for disaster response[13] or installation in construction works[14] – the high-level commands are provided by a human, either via tele-operation[13] or by a pre-defined task sequence[14]. In this paradigm, the autonomous execution of a task builds and relies on the use of planning explicit task sequences, which uses limited sensory feedback to replan online. Such an approach is thus not yet fully autonomous and is vulnerable to uncertainties when the environment is likely to change during the interaction.

For planning methods, such as trajectory optimisation[15] or task planning[16-18] a model of the environment is needed to optimise a motion sequence. Such a planning framework passes commands in a top-down approach and has a separation between planned motions and the control of their executions. Consequently, the planning framework unilaterally connects with the control and lacks having feedback from the low-level layer[13]. Therefore, this approach is restricted to a limited range of scenarios, where the required execution is close to the ideal planning, e.g., quasi-static, kinematic motions or well-defined environments. However, for situations where the environment model deviates from the real world, feedback is indispensable and is required for corrective actions to counterbalance changes that are not planned beforehand. Since the control is responsible for execution and interaction with the environment, the lack of feedback from the lower control layers prevents a wider generalisation to other environments, and hence limits the applicability with respect to autonomous behaviours and human-level motor control.

To overcome the aforementioned limitations, learning approaches – such as Hierarchical Reinforcement Learning [19,20] – is an alternative approach to accomplish tasks that require solving a discrete sequence of sub-tasks in a close-loop fashion, such as path-following for quadruped locomotion[21], interactive navigation with mobile manipulators[22], and hierarchical navigation tasks[19,20] . The remaining challenges[20] lie inter alia in finding the right levels of abstraction, and how to find a proper hierarchical structure with meaningful sub-behaviours. This is especially challenging for robotics, where the state and action space is complex, and behaviours are abstract and often hard be quantified explicitly. Hence, hand-crafted sub-behaviours, such as the theory of *options* used in Sutton et al[23], prevent adequate exploration during reinforcement learning which is needed for autonomous operations.

This study investigates and presents five core principles of human motor control, based on which we formalise the design of a framework that generates autonomous behaviours. Our proposed hierarchical generative model adheres to the core principles of hierarchical motor control[24], and the resulted capability can tackle several challenges that HRL has not yet overcome. Compared to HRL, our ensuing hierarchical control structure offers the possibilities to (i) create a transparent and flexible approach to interpret and implement robotic decision making, (ii) rollout individual policies inside the hierarchical structure and improve their overall performance, and (iii) identify and mitigate the cause of performance deficits.

This work achieves human-level motor control by pursuing the notion that structural dependencies – c.f., inter-region communication as observed in human motor control[24] – are necessary for autonomous robotic systems to optimise and adapt future actions in uncertain environments. Human motor control is generated through nested hierarchies comprising distinct, but functionally interdependent, processing structures, e.g., from the motor cortex to the spinal cord down to neuromuscular junctions[25, 26]. These nested hierarchies can be interpreted as a hierarchical generative (or forward) model[27, 28, 24, 29].

Our model is a particular instantiation of such hierarchical motor control models, as contrasted by the prior studies[30, 31, 12, 32, 33]. To develop further, our extension has introduced and incorporated multi-level planning, (asymmetric) inter-region communication and temporal abstraction analogous into the computational models of human motor control[24, 34, 35, 36, 37].

We demonstrate that these hierarchical generative models contribute to new abilities to perform physical tasks autonomously, in a context sensitive and robust fashion. To this end, we employed a hierarchical generative model, which combines multiple levels of spatial and temporal abstractions and uses the existing state-of-the-art robotics tools, e.g., impedance control, model predictive control, and reinforcement learning.

In this work, we characterise motor control as an outcome of a (learnt) hierarchical generative model; in particular, generative models that include the consequences of action. This proposal inherits from hierarchical functional organisation of human motor control and ensuing planning as inference[38, 39, 33], active inference[40, 41, 32, 42, 43, 44] or control as inference[45, 46, 47,48]. Briefly, hierarchical generative models are a description of how sensory observations are generated, i.e., encodings of sensorimotor relationships relevant for motor control[49,50]. Importantly, this gives for free the five core principles of hierarchical motor control introduced in Merel et al., 2019[24] – see Table 1 for further details.

This hierarchical formulation can facilitate multi-level planning that operates at different levels of temporal and spatial abstraction[24, 50] (see Figure 1). This follows from functional integration of separate planning (i.e., choosing the next appropriate actions), motor generation (i.e., executing the selected actions) and control (i.e., realising high-level plans as motor movements), as provided by the hierarchical generative model[50, 29,51]. As a result, the requisite architecture can be considered as a series of distinct levels, where each provides appropriate motor control[52] (see Figure 1). In our construction, the lowest level predicts the proprioceptive signals—generated using a forward model of the mechanics—and the kinetics that undergirds motor execution. This kinetics can be regarded as realising and equilibrium position or desired set point, without explicit modelling of task dynamics (c.f., the equilibrium point hypothesis[*])[53,54]. The level above generates the necessary sequence of fixed points, that are realised by the lower level. This sequence speaks to the stability control that a human has over limbs, to perambulate in an upright manner over, say, a centre of gravity. The highest level then pertains to planning[29, 55], and different states represent endpoints of an agent's plan, e.g., move a box from a table to another.

To validate our proposition, we introduce a hierarchal generative model for autonomous robotic operations using a learning-based scheme. Our model has three distinct levels for planning and motor generation, emulating a (simplified) functional architecture of human motor control. Importantly, each level comprises separate but functionally integrated modules, which have partial autonomy supported by (asymmetric) inter-regional communication[56] i.e., the lower levels can independently perform fast movements. Such a structure provides a flexible, scaled-up construction of a hierarchical generative model, using established robotic tools (see Methods). The ensuing levels in the model hierarchy were optimised sequentially and evolved at different temporal scales. However, only the middle level planner had the access to state feedback, which allowed for a particular type of factorisation (i.e., functional specialisation) in our generative model. We reserve further details in the later sections.

To show the effectiveness of using hierarchical generative modelling to design fully autonomous and robust robotic systems, we focused on hierarchical task planning that involves the decision-making of distinct sub-tasks to complete the whole mission. The first task required the robot to carry a box from one table to another. Here, the planner needed to prescribe the sequence of (i) approaching and picking up the box, (ii) walking towards the second table, and (iii) placing the box upon arrival. In the second task, the robot needed to reach a goal location by opening a closed door, which only opens after pressing a button. Here, the hierarchical framework needed to coordinate both its legs and arms to (i) walk towards the door, (ii) move arms to press the button on the door, and (iii) pass through the door and reach the goal location.

---

[*] Briefly, the equilibrium point hypothesis states that all movements are generated by the nervous system through a gradual transition of equilibrium points along a desired trajectory.

In both scenarios, our algorithm performed coherent locomotion, manipulation, and grasping movements that are comparable to humans, and successfully solve these complex tasks in a simulation environment. Furthermore, we demonstrated the robustness in presence of uncertainties, such as new environmental changes, unanticipated external pushes, and even the amputation of the right foot. To demonstrate the generality and effectiveness of our proposed hierarchical generative model, we showcase the ability of our method to (1) execute a penalty kick by approaching a ball and striking it towards a goal, and (2) perform a pick-and-place task by retrieving a box from one conveyor belt, transporting it to a second conveyor belt, and activating the second conveyor belt's movement-button to send the box away.

The paper is organised as follows. In Section 2, we demonstrate that our (implicit) hierarchical generative model for motor control, which entails a bidirectional propagation of information between different levels of the generative model, can perform tasks remarkably similar to humans. Here, an implicit hierarchical generative model refers to a forward model whose explicit inversion corresponds to control as inference (without the need for an inverse model). In Section 4, we discuss the effectiveness of our hierarchical generative model, how it may benefit potential applications, and provide an outlook for future work. Lastly, in Section 5, we provide details of our implementation.

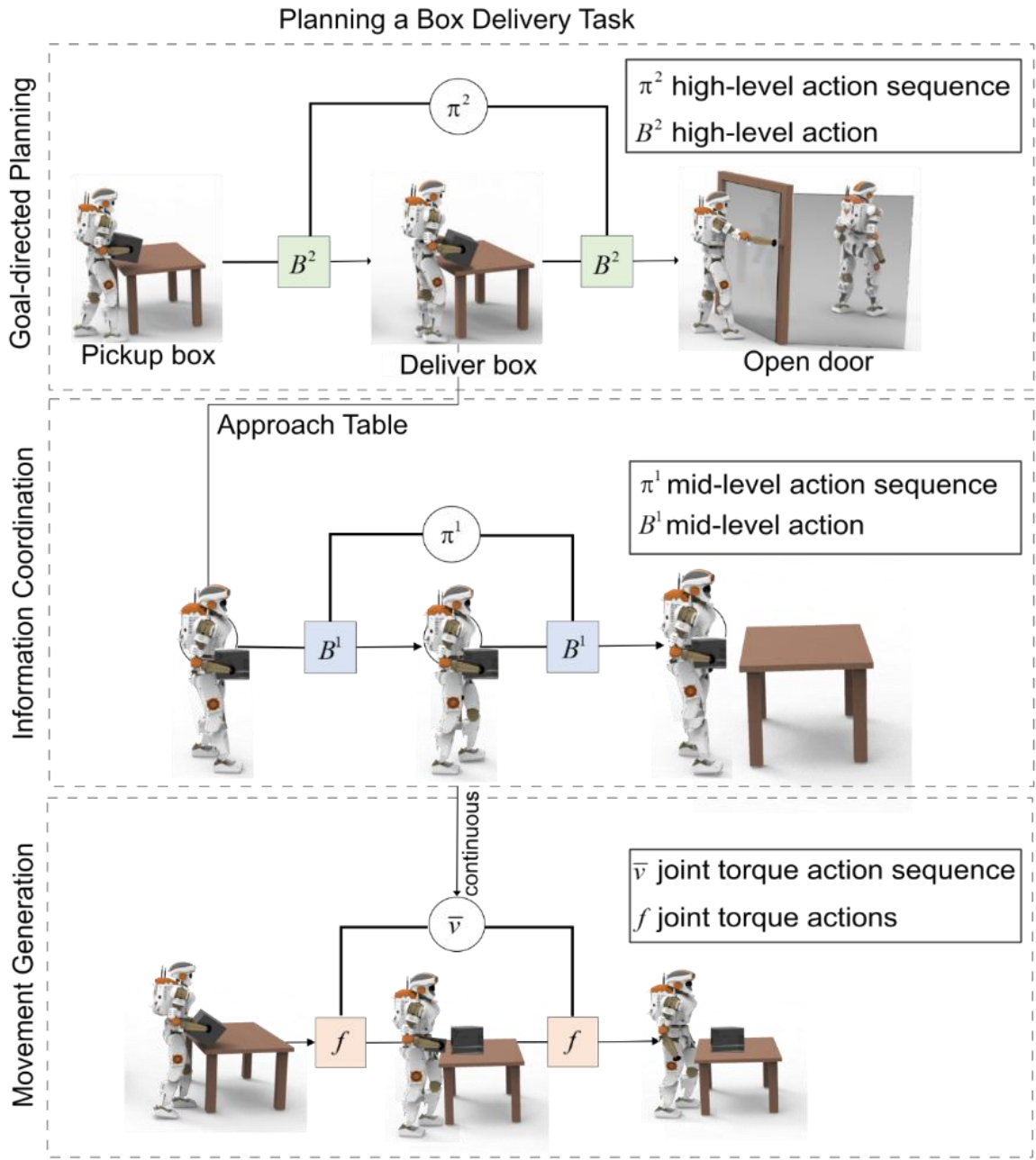

**Figure 1: Pictorial representation of a hierarchical generative model for moving boxes – a form of motor control**. A generative model represents the conditional dependencies between states and how they cause outcomes. For simplicity, we express this as filled squares that denote actions, and circles that represent action sequences. The key aspect of this model is its hierarchical structure that represents sequences of action over time. Here, actions at higher levels generate the initial actions for lower levels– that then unfold to generate a sequence of actions: c.f., associative chaining. Crucially, lower levels cycle over a sequence for each transition of the level above. It is this scheduling that endows the model with a deep temporal structure. Particularly, planning (first row; highest level) to "deliver the box" generates the actions for the information coordination level (second row; middle level) i.e., "movement towards the table". This in turn determines the initial actions for movement generation (third row; lowest level) of arms to "place the box" on the table. Here, a single action is generated at each time step by sampling from action sequences (i.e., sequential policies) that are generated up to a specified time horizon.

**Table 1.** Summary of the key principles of hierarchical motor control[24], with exemplar realisations in human motor control and our robotic system. We omit the principle of modular objectives here (sub-systems trained to optimise specific objectives distinct from the global task objective) because a factorised generative model architecture leads to distinct factor specific objectives at each level in the hierarchy.

| Principle | Description | Hierarchical generative models | Human motor control | Our robotics system for autonomous operations |
|---|---|---|---|---|
| Information factorisation | Different information is processed by distinct sub-systems. | Factorised distribution of appropriate latent states within the generative model. | Different sensory signals are routed to different parts in the hierarchy, e.g., what and where streams. These neuronal pathways can be characterised as factorised states responsible for sub-systems. | Only task-relevant sensory signals are used by individual levels, with irrelevant states hidden across levels. This speaks to an *explicit factorisation of sensory signals* and which parts of the system have access to them. |
| Partial autonomy | Lower hierarchical levels can semi-autonomously produce outputs with minimum input from levels above. | The result of factorising state space into multiple levels can independently accomplish sub-goals at a (relatively) fast temporal scale. | Semi-autonomous coordination of joint movement at lower levels (i.e., brainstem and spinal cord). These operate at a faster temporal scale and do not require continuous input for higher levels. | Full autonomy and stability guaranteed at individual levels. Explicitly, we introduce stable mid-level and low-level motions for random higher-level inputs. This ensures that lower levels can independently perform fast movements. |
| Amortised control | Re-execute appropriate behaviours rapidly using learnt movements. | Learnt probability distributions that parameterise this generative model can be used for amortised control. That allows for habitual control based on previously learnt distributions. | The cerebellum is responsible for amortised control of deliberative and goal-directed behaviours, evoking fast habitual control for repeated actions. | The system learnt policies (i.e., action-state mappings) that provide habitual control for rapidly re-executing appropriate actions. |
| Multi-joint coordination | Degenerate coupling of different components operating as a whole for motor control. | Result of state factorisations that introduce flexible mapping across and within each level. | Different neuronal ensembles have distinct influences e.g., the red nucleus controls movements of the arms. Much like factorised states, these neuronal ensembles come together to produce intricate movements. | The system is equipped with multiple sub-structures (or policy mappings) that are responsible for specific actuator movement. Together these come across, and within levels, to produce particular motor movements. |
| Temporal abstraction | Abstraction of time across hierarchical levels. | A feature of hierarchical generative models, where higher levels evolve slower than and constrain, the level below. | Different levels evolve at different temporal and spatial scales, with the primary motor cortex responsible for planning (slow timescale) and spinal cord responsible for generation (fast timescale) | The three levels of the system evolve at different temporal scales – much like any hierarchical generative model. The high-level planning is at a slow timescale, mid-level stability control at medium timescale, and low-level joint control at a fast timescale. |

# RESULTS

Our implicit hierarchical generative model enables a robot to learn how to complete a loco-manipulation task autonomously in simulation. We validate this model in three distinct scenarios: (1) a sequential task with two-step decision-making that involves moving a box from one table to another and opening a door by pressing a button (Fig. 1); (2) transporting a box between conveyor belts and activating the second belt by pushing a button (Fig. 2a); and (3) executing a penalty kick by approaching and kicking a football into a goal (Fig. 2b). The learned policy demonstrates generality and robustness to uncertainty (Fig. 3a-e), while evincing the core principles of hierarchical motor control.

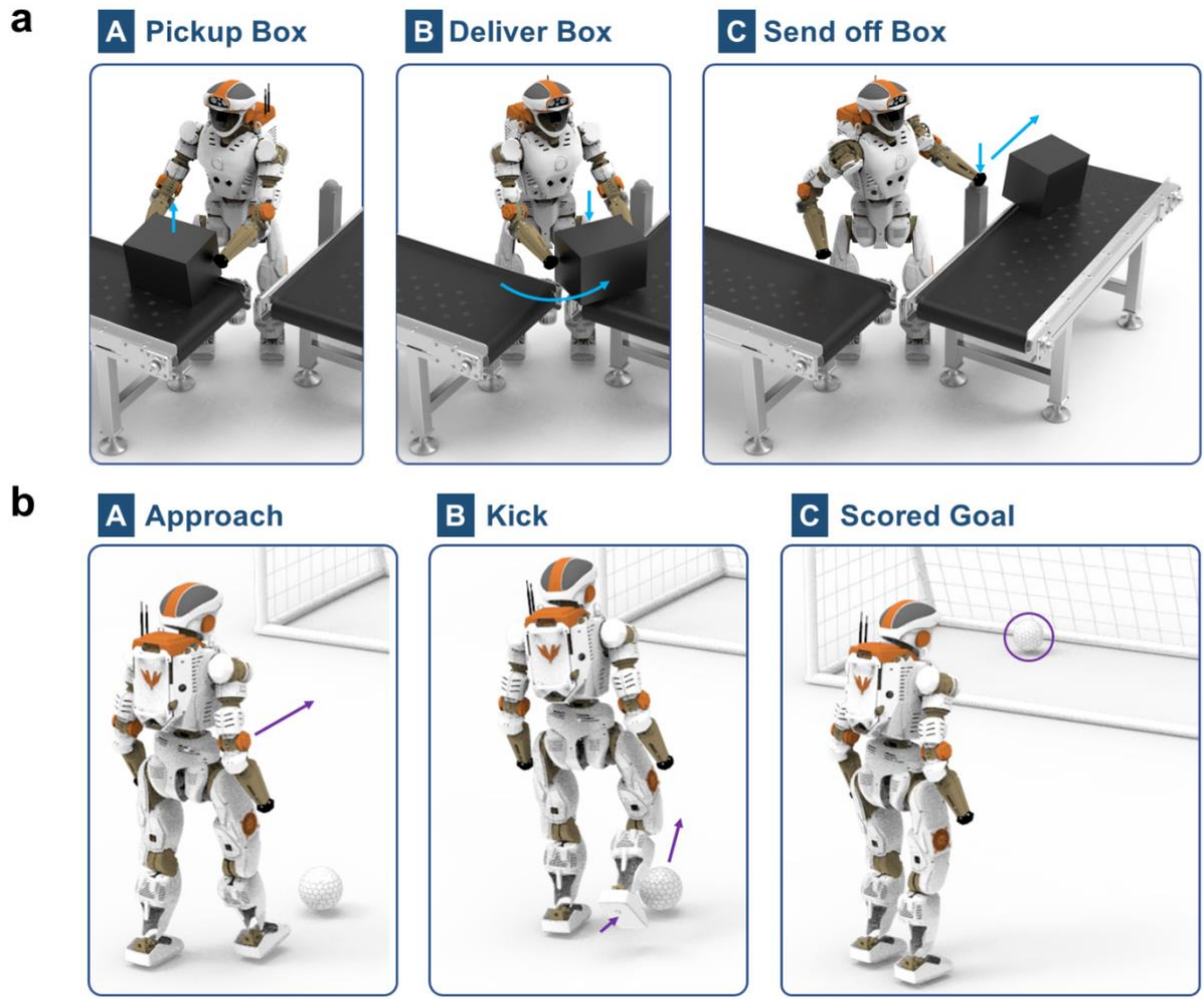

**Figure 2:** Manipulation and locomotion tasks to validate the hierarchical generative model. Panel a illustrates a manipulation task, where the robot picks up the box, delivers it, and finally sends it off by activating the button. Panel b shows a penalty kick, where the robot approaches the ball and kicks it into the goal.

Our implicit hierarchical generative model can successfully and autonomously achieve locomotion, manipulation, and grasping movements like humans, and solve all these complex tasks coherently with internal consistency. Briefly, the highest policy level of the robot system determines the action sequence necessary for task completion and sends commands to the lower levels. This allows the robot to carry out the following actions: (i) walk to the first table, (ii) move arms to pick up the box, (iii) walk to the second table, (iv) and place the box on the table. Upon successful completion of this task, the same generative model can be used to perform the second task as well. This has been achieved by using the high-level policy to open a closed door, instead of moving a box. As the controller is formulated as part of a hierarchical generative model, task adaption reflects high-level policy learning, because only the high-level policy has access to task-relevant states, e.g., position of robot with respect to box, door, etc.

The accompanying commands from the high-level policy are sent to the lower levels responsible for limb stability and joint control. They instantiate mid-level locomotion and manipulation policies to move to the door and position the body close enough to press and opening button. Having opened the door, the robot enters, and proceeds towards the final goal. Here, independently, the locomotion policy can facilitate adaptation to perturbations, e.g., recovering from pushes or locomotion over different types of terrains. Contrariwise, the low-level joint controller provides robustness to sudden and hard contacts with the ground. This is by design, where the Impedance Controller is optimised to absorb high-frequent impacts.

To assess the robustness and generality of this hierarchical scheme, we introduced several perturbations that were not encountered during training (Fig. 3). First, we introduced external perturbation by placing obstacles (i.e., 5kg box, Fig. 3a) in front of the robot and pushing its pelvis (Fig. 3b). The results were encouraging: the mid-level locomotion policy withstood both perturbations, moved the obstacle out of the way, and took a step to recover balance after the push. To test the performance further, we modified the environment with unseen conditions by adding a 5° inclined surface (Fig. 3c) and a low-friction glass plate (friction coefficient of 0.3, Fig. 3d) in front of the door. The robot could complete the task after each perturbation. More interestingly, we lesioned the robot by amputating its right foot (Fig. 3e). Despite this handicap (that was never encountered)—and with only a stump touching the ground in place of its right foot—our hierarchical control was sufficiently robust to deal with this situation and the robot was able to keep balance and complete the task.

Next, we evaluate whether the ensuing control architecture satisfies the key principles of hierarchical motor control (Table 1) that underwrite robust task performance.

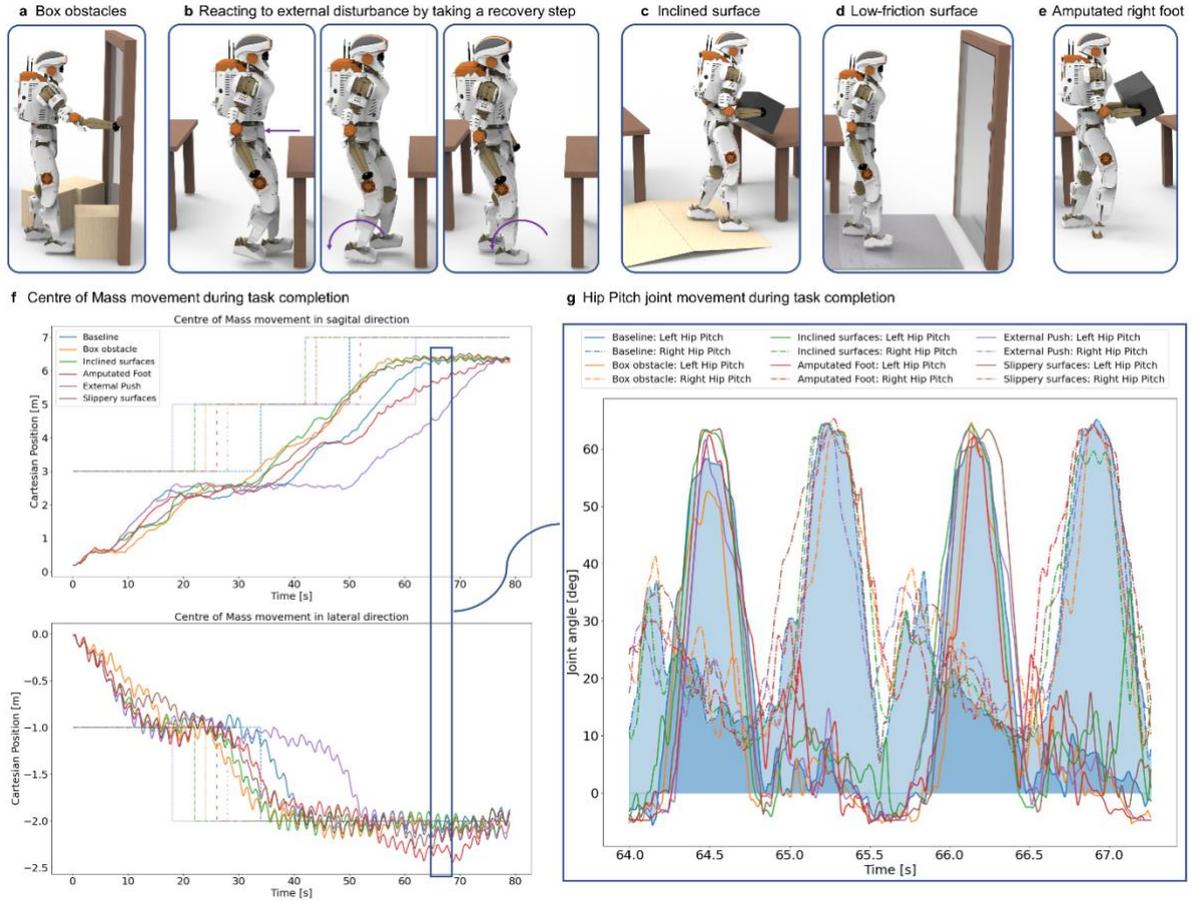

**Figure 3:** Robustness of the system in the presence of perturbations and environmental changes. Panels **a-e** show how the robot completes the task in perturbation test scenarios that it has not encountered during training and demonstrate the robustness of our proposed method. From left to right, we place 5kg box-obstacles in front of the robot, push the robot from the front, alter the floor with an inclined and slippery surface, and lesion the robot, but removing the right foot. In Panel **f**, the sagittal and lateral Centre of Mass movement is shown under different perturbations demonstrating the amortised control. Panel **g** shows the hip pitch joint movement, which has the biggest effect on the motion during biped locomotion). The hip pitch joint motion is used to show how the policy adapts to the perturbation and rapidly re-executes a motion to counteract the perturbation.

## INFORMATION FACTORISATION

In this system, factorisation exists across model levels and policy controls, each responsible for a particular sort of information processing. This factorisation ensures that external perturbations have minimum impact on task performance. For example, when disturbances are applied on the robot's legs, the resulted perturbations do not significantly impede task completion. This is because the perturbations are only visible to the locomotion policy, and other levels and control policies are conditionally independent of the disturbances and hence retain their nominal operations.

Since the information factorisation defines the role for each sub-system, thus, any failures in performance can be isolated and fine-tuned for future tasks. For example, if the robot falls over while walking to a goal, the locomotion policy can be identified as the root cause, and hence improving the locomotion policy will resolve the issue without needing to modify the high-level planner or the manipulation policy. Further examples include oscillation of the robot limbs, which can be attributed to the low-level joint control; or walking in the wrong direction, which was due to the command from the high-level policy. From a theoretical perspective, but not implemented in our hierarchical generative model, factorisation of this sort corresponds to the structure of the generative model that can be decomposed into factors of a probability distribution (in physics and probabilistic inference, this is called a mean field approximation). Almost universally, this results in certain conditional independencies that minimise the complexity of model inversion; namely, planning as inference or control as inference[38, 33, 41,57]. This is important because it precludes over fitting and ensures generalisation. From a biological perspective, this kind of factorisation can be regarded as a functional segregation that is often associated with modular architectures and functional specialisation in the brain[56].

### PARTIAL AUTONOMY

The system is designed with partial autonomy, i.e., minimum interference or support from other levels. Specifically, we implement a clear separation between the highest and intermediate levels, though they are learned together. This is particularly relevant because the high-level planning level could send unrealisable action sequences to the mid-level stability controller. Without partial autonomy, the robot can become unstable and unable to learn to move appropriately, given such random or potentially unstable high-level commands.

Figure 4 illustrates a case when the robot is provided with random commands to both the arms and legs. This causes the robot to walk in random directions (Fig. 4a) and the arms move around randomly (Fig. 4b). Despite imperfect motion tracking, the robot does not fall over and can complete the tasks despite incoherent intentions.

**a** Random leg motion: Sagital CoM motion

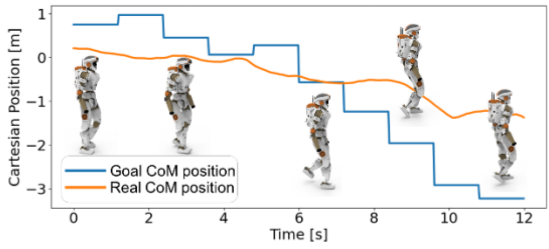

**b** Random arm motion: Sagital CoM motion

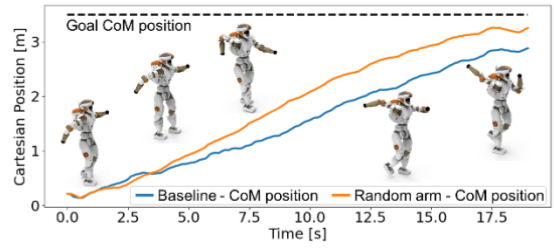

**c** Leg joint movements

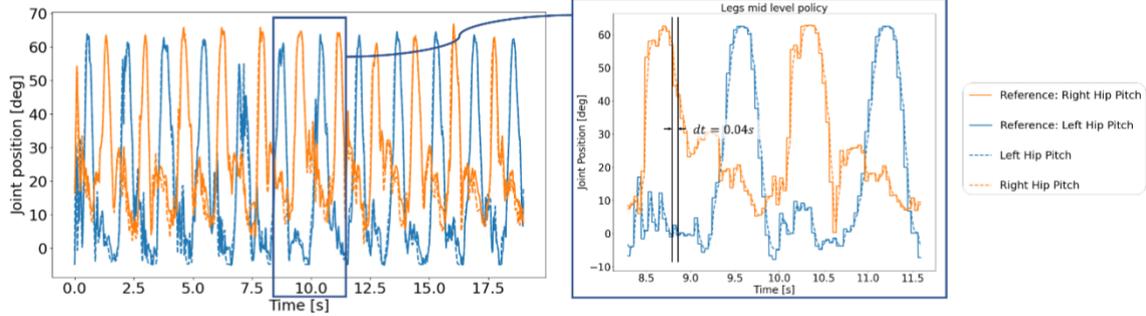

**d** Arm joint movements

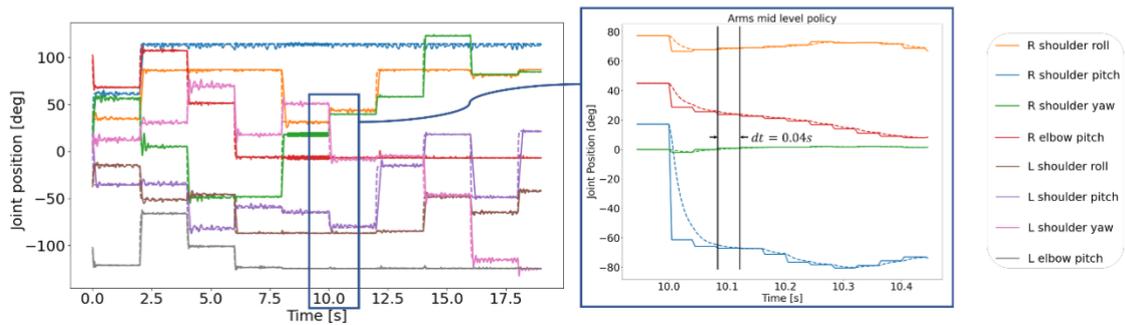

**Figure 4:** State and temporal dynamics of the robot during task performance with random-high level commands. Panels **a** and **b** show the sagittal motion of the Centre of Mass (CoM) while following random leg and arm commands respectively. From the robot snapshots corresponding to the time they're shown, the partial autonomy of the mid-level stability controllers can be seen, i.e., good performance of the individual levels despite random and fast-changing command inputs. Panels **c** and **d** show the leg and arm movements respectively. Here, the separation of temporal scales during planning can be seen, where the high-level commands are provided at 0.5 Hz, and the mid-level commands are realised at 25Hz. The joint commands are realised at 500Hz on the joint actuators. In the inset plots of Panels **c** and **d**, the joint position trajectories evolve similarly as postulated in the equilibrium point hypothesis.

## AMORTISED CONTROL

After training, the robot engages in amortised control with the ability to re-execute appropriate behaviours rapidly using previously learnt movements. We observed this behaviour in the baseline and perturbed task settings (inset trajectories in Fig. 3f), where the amortised locomotion policy was used to complete the task without the need of additional learning.

## MULTI-JOINT COORDINATION

The robot has multiple sub-structures that are responsible for specific controls and work together in different ways to generate motor movements. Figure S5a demonstrates this multi-joint coordination when pressing the button to open the door in the presence of an obstacle (Task 2). To achieve this, the right arm motions had to coordinate appropriately according to the initial hand position. Also, the shoulder roll (Fig. S5b orange line) and elbow (Fig. S5b red line) had to adjust and adapt differently from the baseline. Explicitly, these do not yield a fixed motion, instead, the manipulation policy coordinates these joints based on the Centre of Mass (CoM). Therefore, during the baseline reaching motion, the arms move differently than that in the case of an obstructed box, where the CoM is in a different position because boxes are obstructing the door.

## TEMPORAL ABSTRACTION AND DEPTH

By design (see Subsection "Implicit Generative Models"), the three system levels evolve at different temporal scales. Figure 4 illustrates these distinct scales as the robot perambulates. The highest policy level has a slow timescale of 0.5Hz (Fig. 4a). This allows the lower levels to carry out the command in a partially autonomous way, i.e., uninterrupted. Conversely, the mid-level stability control of limbs has a faster timescale at 25Hz (see the inset trajectories of Fig. 4c and 4d). This is needed to generate rapid predictions for the locomotion and manipulation policies. Finally, the low-level joint control executes these control commands at a frequency of 500Hz on the actuator level.

# DISCUSSION

## HIERARCHICAL GENERATIVE MODELS OF MOTOR CONTROL

Our hierarchical generative model is an abstract computational representation of the functional architecture of human motor control (Figure 5). Here, we briefly discuss its computational neuronal homologues, focusing on predictions of primary afferent signals from muscles, and consider the corresponding principles for human motor control. The inversion of forward models—that underwrite human motor control—generates continuous proprioceptive predictions at the lowest level and propagates information to the highest levels that are responsible for planning. Accordingly, our formulation provides an implicit generative model that can be used by a model-based robotic agent, including reinforcement learning and active inference[58], in order to infer its environment dynamics.

The generative model's lowest and fastest level includes the spinal cord and the brainstem. These areas are responsible for evaluating the discrepancy between the proprioceptive inputs (primary afferents) and descending predictions of these signals. This discrepancy (a.k.a. prediction error) drives the muscle contraction via classical motor reflexes and their accompanying musculoskeletal mechanics[29,59]. On this view, descending predictions of proprioceptive input supply equilibrium or setpoints that are fulfilled by classical reflexes [60 , 61, 62]. This is instantiated in our model at the low-level joint control, which receives current joint position and sensor information to calculate the desired torque necessary for achieving a targeted and predicted position (supplied by the mid-level controller) via motor control. Here, the joint controller has partial autonomy to compute the desired torque, similar to neuronal ensembles (i.e., the red nucleus) controlling low-level arm movements.

At an intermediate level, one could consider the role of the cerebellum. The cerebellum receives ascending inputs from the spinal cord, and other areas, and integrates these to fine-tune motor activity. In other words, it does not initiate movement, but contributes to its coordination, precision, and speed, through a fast non-deliberative mode of operation. Therefore, it can be thought of as being responsible for amortised (habitual) control of motor behaviour, which is characterised by subcortical and cortical interactions[63, 64, 65, 66, 67]. The cerebellum receives information from the motor cortex, processes this information, and sends motor impulses to skeletal muscles (via the spinal cord). The mid-level of our generative model is used for similar coordination and stability control of locomotion and manipulation policies that yield multi-joint coordination. It fine-tunes pelvis and hand targets, given descending policy from the higher level, to determine exact joint location (measured in radians). Like the cerebellum[68, 69,] this level can coordinate multiple joint movements semi-autonomously over time.

Higher levels of the generative model include the cerebral cortex, among other neuronal systems. The cortex has access to factorised sensory streams of exteroceptive, interoceptive and proprioceptive signals (e.g., visual, auditory, somatosensory, etc) and can coordinate, contextualise, or override habitual control elaborated in lower levels. Specifically, the primary motor cortex is responsible for deliberative planning, control, and execution of voluntary movements: for example, when learning a new motor skill prior to its habituation or amortisation.

These are instantiated as ascending tracts that cross over to the opposite side of the system, e.g., the spinocerebellar tract that is responsible for sending sensory signals regarding arms and limb movements. Conversely, descending tracts carry appropriate motor information to the lower levels e.g., the pyramidal tracts responsible for sending conscious muscle movements. The role of the cortex is instantiated at the highest level of our model, with access to processed sensor information to aid decision making. Specifically, we introduce asymmetric inter-region connections with connections from the low-level sensor information to this high-level, and from this high-level to the mid-level stability control. Anatomically, these correspond to extrinsic white-matter connections in the brain which, in predictive coding and variational message passing schemes, are responsible for belief updating and planning as inference[50,61].

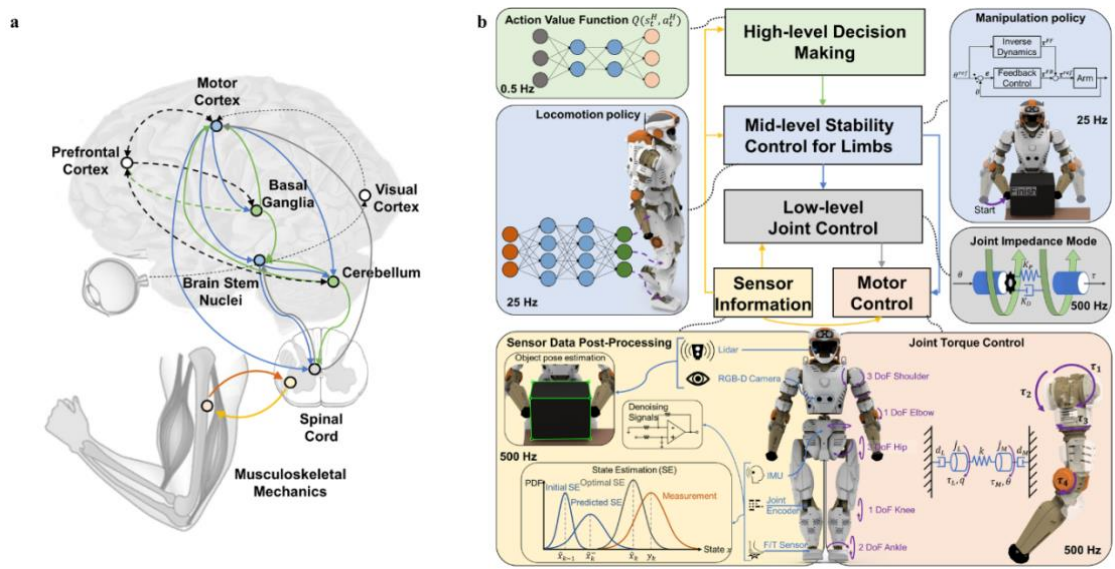

**Figure 5:** Algorithmic realisations of hierarchical control as inference. Panel **a** presents a schematic of a (high-level) generative model that underwrites human motor control, and Panel **b** depicts the implicit generative model for a robotics system. The green nodes in Panel **a** and green boxes in Panel **b** refer to the highest levels of human motor control and our implicit generative model respectively. In the generative model, high-level decision making is realised as a neural network learned through Deep Reinforcement Learning. The blue nodes in Panel A correspond to the middle level of human motor control and the blue boxes in Panel B are intermediate level realisations, implemented as Deep Neural Network policy learned through Deep Reinforcement Learning for Locomotion and Inverse Kinematics and Dynamics policy for Manipulation. On the lowest level, depicted in grey nodes and boxes, a joint impedance controller calculates the torques required for the actuation of the robot. Yellow and light red denote sensor information and motor control respectively. For clarity, we limit our exposition to key regions in Panel **a**, based on prior literature, where these are drawn using the solid lines. The dotted lines represent the processing of a separate outcome modality for human motor control, i.e., the visual input. Lastly, the prefrontal cortex is connected via the dashed lines to denote its supporting role during human motor control. Dotted lines in Panel **b** indicate the realisations of the corresponding principles, while dashed lines indicated message parsing. Please refer to subsection "Implicit hierarchical generative model for a robotics system" for the algorithmic implementation of Panel **b**.

**FUTURE DIRECTIONS**

By providing robots with a new level of task autonomy for both locomotion and manipulation skills—with appropriate triage procedures—humans can be relieved from the necessity of sending low-level commands for control and decisions to robots, e.g., foot and hand contacts, as commonly seen in a shared autonomy and semi-autonomous paradigms. Consequently, we can overcome potential limitations coming from human errors and the reliance on the communication bandwidth. One example is the large number of robots that fell during the DARPA Robotics Challenge Finals in 2015[70], where robots had very little autonomy and relied on close supervision by humans, such that the whole scheme became error-prone and vulnerable, which suffered from erroneous human decision-making, lack of local robot autonomy against environmental uncertainties and disturbances, and so on.

Future work will evaluate the use of hierarchical generative models under more nuanced planning objectives, and different autonomous robotics systems. Because of the modular factorisation of the implicit hierarchical generative model, policies at various levels can be replaced and further upgraded with an alternative controller or a learned policy. For example, replacing our Q-learning planner with more sophisticated schemes which are designed to handle aleatoric and epistemic uncertainties (i.e., expected free energy[71, 72, 73]). This type of future work can improve the performance in volatile conditions[72].

Furthermore, robustness can be evaluated through robotic neuropsychology[74], i.e., introducing in-silico lesions by perturbing various approximations and policies and investigating their effect on the ensuing inference and behaviour. These computational lesions can be introduced in both simulated and physical robots, where lesions of this sort can change functional outcomes. For example, perturbations on the minimum-jerk optimisation solution (i.e., computational lesion) at the mid-level stability control would lead to cerebellar tremors for the arms.

# METHODS

Here, we present the hardware implementation for inverting the (implicit) hierarchical generative (a.k.a. forward) model for autonomous robot control. The specification of the robot platform can be found in the Supplementary Material. First, we detail the task that is completed autonomously by inverting the generative model, i.e., using the model to predict sensor inputs and using actuators to resolve the ensuing (proprioceptive) prediction errors. Next, we elaborate on the details of generative model including high-level decision making, mid-level stability control, and low-level joint control.

## TASKS OF INTEREST

To demonstrate how in inversion of—or inference under—a hierarchical generative model solves complex tasks that require a particular sequence and coordination of locomotion and manipulation skills, we designed a task that demanded both coordination of limbs and reasoning about the sequence of actions. This task comprised four subtasks (Fig. S1): picking up a box from the first table, delivering the box to the second table, opening the door, and walking to the destination or goal position. To complete the task, all the subtasks had to be carried out in an exact sequence.

Our proposed framework allowed the robot to learn successful task completion through interactions with the environment in simulation. This was achieved by designing a reward (or utility) function for the high-level policy, such that cumulative maximisation of reward leads to task completion (see

Section "High-level Decision Making"). For the mid- and low-level policies, a combination of control policies and imitation learning was used.

The trained policy learnt to first pick up the box, and hold the box using its hands, while approaching the second table. Once the box could be safely placed at the second table, the selected policy ensured the box was released and the robot moved towards the door, while keeping its arms in an appropriate position. Next, by pushing the button placed on the right side of the door, the robot could open the door. The learnt policy ensured that the destination was approached at the right time, i.e., when the door was sufficiently open to walk through—towards the goal position.

## IMPLICIT HIERARCHICAL GENERATIVE MODEL FOR A ROBOTICS SYSTEM

Following the key principles of hierarchical motor control in Table 1 and the generative model in Figure 1, we constructed a generative model for a humanoid robot comprising three levels: high-level decision making, mid-level stability control, and low-level joint control. The structure of the hierarchical generative model is shown in Figure 5b. This hierarchical architecture rests on conditional independencies that result in factorised message passing between hierarchical levels[†]. Here, the temporal depth and structure of motor planning rests on specifying a hierarchical generative model, where level-specific policies are evaluated at different timescales. In this setting, each level assimilates[75] evidence from the level below, in a way that is contextualised or selected by (slow) constraints, afforded by the level above. A summary of the implicit hierarchical generative model for a robotics system can be seen in Table S4 and Figure S4.

The (implicit) hierarchical generative model is instantiated as:

---

[†] It shall be noted that the implementation used in our simulations is not formulated explicitly in terms of message passing or belief updating, which are usually articulated in terms of Bayesian filtering for continuous states and belief propagation, or variational message passing for discrete states. However, there exists an interpretation of the scheme, in terms of what expected states of the environment that can cause such sensor inputs. Crucially, most of these causes correspond to the action of the robot itself. In our context, factorised propagation of messages is a consequence of routing of relevant information to different levels of the hierarchy. For example, the mid-level locomotion policy only has access to the lower joint states and goals, i.e., desired pelvis position and proprioceptive feedback states of the robot.

$$p\left(o_{0:T_1}, s_{0:T_n}^{1:3}, a_{0:T_n-1}^{1:3}\right)$$

$$= \underbrace{\Pi_{n=1}^{N} p(s_0^n) p(a_0^n)}_{\text{state \& action prior}}$$

$$\times \Pi_{t_3=1}^{T_3} \Pi_{t_2=1}^{T_2} \Pi_{t_1=1}^{T_1} \left[ \underbrace{p\left(s_{t_3}^3 \middle| s_{t_3-1}^3, a_{t_3-1}^3, o_{t_1-1}\right)}_{\text{level 3 transitions}} \underbrace{p\left(a_{t_3}^3 \middle| s_{t_3-1}^3\right)}_{\text{level 3 policy}} \right.$$

$$\times \underbrace{p\left(s_{t_2}^2 \middle| s_{t_2-1}^2, a_{t_2-1}^2, o_{t_1-1}, a_{t_3}^3\right)}_{\text{level 2 transitions}} \underbrace{p\left(a_{t_2}^2 \middle| s_{t_2-1}^2\right)}_{\text{level 2 policy}}$$

$$\left. \times \underbrace{p\left(s_{t_1}^1 \middle| s_{t_1-1}^1, a_{t_1-1}^1, o_{t_1-1}, a_{t_2}^2\right)}_{\text{level 1 transitions}} \underbrace{p\left(o_{t_1} \middle| s_{t_1}^1\right)}_{\text{likelihood}} \underbrace{p\left(a_{t_1}^1 \middle| s_{t_1-1}^1\right)}_{\text{level 1 policy}} \right],$$

where outcome $o_t \in O$, state $s_{t_n} \in S$, action $a_{t_n} \in A$, and $p$ denotes a probability distribution. The superscript $n \in \{1,2,3\}$ indicates the level of the state $s^n$ or action $a^n$, with $N = 3$ being the highest level and $n = 1$ being the lowest level. The subscript $t_n \in \{1, \dots, T_n\}$ indicates the time at each level $n$ evolving at different temporal scales: the highest level ($n = 3$) evolves at 0.5 Hz, the mid-level ($n = 2$) at 25 Hz, and the lowest level ($n = 1$) at 500 Hz[‡]. This temporal ordering denotes how different levels contextualise the level below: the high-level policy contextualises the roll-out for the mid-level; mid-level policy contextualises the low-level; and each level has access to previous outcomes. Briefly, the transition function is defined as an identity – using the previous outcome – for levels 2 and 3 until the next update (i.e., 50 level 1 steps for level 2, and 1000 level 1 steps for level 3). The pseudocode for optimising each level can be found in Figure S4 of the Supplementary Material, along with a detailed overview of dependencies across levels.

The highest planning level, evolving at the slowest rate, selects an appropriate sequence of limb movements, which are needed to complete a particular sub-task. It decides where the hands should be and what direction to go. Practically, deep reinforcement learning (RL) is used to learn a high-level decision-making policy that generates targets (in a Cartesian space) for the mid-level stability control: c.f., the equilibrium point hypothesis for human motor control[60] and active inference formulations of oculomotor control[50].

---

[‡] To implement nested timescales (i.e., rollout of different temporal scales at each hierarchical level) in the physics simulation, we employ the step function in Pybullet physics simulator to step and wait in the simulated environment. The number of ticks to wait is determined by the ratio of the lower level's frequency divided by the higher level's frequency. As for the real hardware implementation, the hierarchical levels can be realized through multiple threading and synchronized via mutexes on real-time Linux kernels, such as RTLinux [76].

These planning targets are realised at the level below that regulates the balance and stability of the robot during manipulation and locomotion. Manipulation is instantiated as a minimum-jerk model-predictive controller that moves the arms to the target positions provided by the high-level policy. Locomotion is implemented as a learnt mid-level policy, via deep RL, that coordinates legs—with twelve degrees of freedom—to reach the destination predicted by the higher level. Both policies are designed to ensure that infeasible set points—from the high level—are corrected for the mid-level stability control so that only stable joint target commands are supplied to the low-level joint controller.

Despite receiving inputs from other levels, each level has partial autonomy over its final predictions and goal. Furthermore, multi-joint coordination is realised by learning a policy that coordinates all joints of legs appropriately for the current state, while the arms coordinate their joints through Inverse Kinematics (IK).

The low-level joint controller is instantiated as joint impedance control and tracks the joint position commands afforded by the mid-level stability controller. Based on tuned stiffness and damping, the joint impedance control calculates the desired torque to attain target positions closely and smoothly. Lastly, the torque commands are tracked by the actuators, using embedded current control of onboard motor drivers.

The components of the hierarchical generative model can be replaced by another controller or learned policy that has similar performance. In general, controllers are preferred for tasks that are well-specified and can thus be robustly designed: as is the case for manipulation tasks, where only kinematics need to be considered. For tasks with complex interactions, dynamics, or narrow stability regions, learned policies can yield better results. For example, in bipedal locomotion, the problem formulation may become prohibitively complex and therefore stable control becomes difficult due to contact switches of the legs and hard-to model dynamics of the robot. A policy learned via DRL can overcome these problems and produce robust locomotion policies[77].

**Training Process**

The generative model was realised by implementing three levels of control in a hierarchical manner (Fig. 5): high-level decision making, mid-level stability control, and low-level joint control. All components were designed and trained separately, starting from the lowest level.

First, accurate and robust motor control needed to be guaranteed, such that the low-level joint position control could be realised. Stiffness and damping parameters were tuned to track the references accurately and compliantly, which provided the mid-level stability control. The mid-level stability control consisted of a manipulation and a locomotion policy, which were individually designed. The locomotion policy was trained to walk towards a commanded goal position, while the manipulation policy was designed to place the hands on a target position. Finally, the high-level decision-making policy was trained via Deep Reinforcement Learning, which learnt to provide appropriate commands to these mid- and low-level policies.

Due to the modular nature of the implicit hierarchical generative model, the design and tuning of the individual components can be conducted separately and thus do not confound the functionality of remaining components. For example, the neural network architecture design of the high-level planner can be revisited separately from the design of the mid-level policies. Therefore, if the hierarchical generative model is generalized to another robot—that uses a different mid-level locomotion policy— the other components can be retained, e.g., replacing the mid-level locomotion policy with a controller.

**High-level Decision Making**

We achieved high-level decision making, the correct sequence and choices of robot actions, through training a Deep Neural Network that approximated the action value function $Q(s, a)$ over the environment and choose the action $a$ which yielded the highest value in state $s$.

We used Double Q-learning[78] to train a Q-network $Q(s, a; \phi)$, parametrised by weights $\phi$, to approximate the true action value function $Q(s, a)$. At run-time, the action $a$ was obtained as the argument of the maximum Q-value $a = \mathrm{argmax}_a Q(s, a; \phi)$ in state $s$. Two separate Q-networks $Q_1, Q_2$ for action selection and value estimation. Having two separate Q-networks has previously shown to improve training stability[78].

The network parameter $\phi_i$ was obtained by $\min_{\phi_i} L(\phi_i)$:

$$\min_{\phi_i} E\left[\left(r + \gamma Q_j\left(s', a^*; \phi_j\right) - Q_i(s, a; \phi_i)\right)^2\right],$$

with reward $r$, discount factor $\gamma$, network parameters $\phi_i, \phi_j$, Q-networks $Q_i, Q_j$, current state $s$, next state $s'$, best action $a^* = \mathrm{argmax} Q_i(s, a; \phi_i)$. During training, either network parameters $\phi_1$ or $\phi_2$ was randomly selected, trained, and used for action selection, while the other network parameter was used to estimate the action value. The tuple $(s, a, r, s') \sim U(D)$ was obtained from the Experience Replay by uniformly sampling from buffer $D$, which was updated by online action rollout. The time horizon of the high-level decision-making system is implicitly specified with the discount factor $\gamma$ that is used to calculate the return as $G_i = \sum_i \gamma \, r_i$. A way to interpret the discount factor with respect to planning horizon is the concept of half-life of the future reward, i.e., when the current reward $r_i$ is entering the return calculation as $\frac{1}{2} r_i$. With the standard discount factor $\gamma = 0.95$ used in this work, the policy looks ahead ~13.5 steps: $\gamma^{steps} = 0.5 = 0.95^{steps} \Rightarrow steps = \frac{log(0.5)}{log(0.95)} \approx 13.5$. At a control frequency of 0.5Hz, the prediction horizon is roughly 27 seconds.

### Task 1: Box Delivery and Opening Door

The high-level policy sent and updated the actions $a^3 \in \mathcal{A}^3 \subseteq \mathcal{R}^9$ at 0.5 Hz frequency, which were the positions in Cartesian space for the pelvis $a^3_{\mathrm{pelvis}} \in \mathcal{R}^3$, left and right hands $a^3_{\mathrm{lh}}, a^3_{\mathrm{rh}} \in \mathcal{R}^3$. These actions $a^3$ were executed by the mid-level stability controller.

The states $s^3 \in \mathcal{S}^3 \subseteq \mathcal{R}^{12}$ were the vector $\vec{s}_{\mathrm{pelvis}} = p_{\mathrm{table}} - p_{\mathrm{Pelvis}} \in \mathcal{R}^3$ from the table (origin of the coordinate system) to the current pelvis position $p_{\mathrm{Pelvis}}$, and the vectors $\vec{s}_{\mathrm{lh}} = p_{\mathrm{box}} - p_{\mathrm{lh}} \in \mathcal{R}^3$, $\vec{s}_{\mathrm{rh}} =$ $p_{\text{box}} - p_{\text{rh}} \in \mathcal{R}^3$ from current hand positions $p_{\text{lh, rh}}$ to box' position. Lastly, three Boolean variables $o^3 \in O^3 \subseteq [0,1]^3$ were provided as the observation state when the door was open, the box was on the table, or the box was being carried.

The reward terms $r_i$ were determined based on the task completion, such as whether the robot had passed the delivery table, the arm joints were in the nominal position, the box was between the robot hands, the box was at the delivery table, the door was open, and whether the robot was at the goal. The weights $w_i, i = 1, \ldots 6$ can be found in Table S2 (top).

At each time step, the reward $r$ was the sum of sparse, Boolean states:

$$r = w_1 r_{\text{pt}} + w_2 r_{\text{jn}} + w_3 r_{\text{bih}} + w_4 r_{\text{bot}} + w_5 r_{\text{do}} + w_6 r_{\text{ag}},$$

with passed table reward $r_{\text{pt}}$, joints nominal reward $r_{\text{jn}}$, box in hand reward $r_{\text{bih}}$, box on table reward $r_{\text{bot}}$, door open reward $r_{\text{do}}$, and at goal reward $r_{\text{ag}}$.

We terminated the episode early if the robot fell, or collided with itself, tables, or the door. By terminating an episode early—when a sub-optimal state (e.g., falling) is reached—the return is lower, and the policy is thus discouraged from entering similar sub-optimal states.

We initialised the robot in different positions in the environment, such as close to the final goal, in front of the door, or at the second table, to allow the robot to encounter such states that were hard to discover merely by exploration, as a particular sequence of actions were required to reach those states.

### Task 2: Penalty Kick

To perform a penalty kick, i.e., approaching and shooting a ball, the high-level policy is trained similarly to Task 1. Cartesian space commands of the pelvis $a_{\text{pelvis}}^3 \in \mathcal{R}^3$ are generated by the high-level policy (actions $a^3 \in \mathcal{A}^3 \subseteq \mathcal{R}^3$) and executed by the mid-level stability controller.

The states $s^3 \in \mathcal{S}^3 \subseteq \mathcal{R}^4$ are the horizontal positions of the ball and pelvis. A reward r =1 is given, whenever the ball surpasses the goal line, i.e., a goal was scored. An episode is terminated early if the robot fell or collided with itself.

***Task 3: Transporting Box and Activating Conveyor Belt***

The box transportation task consists of two separate sub-tasks that need to be performed in a specific sequence: grasping a box from the first conveyor belt, transporting the box to a second conveyor belt by rotating the torso around the yaw axis, dropping the box off, and sending it away on the conveyor belt by activating the button.

The action space ($a^3 \in \mathcal{A}^3 \subseteq \mathcal{R}^7$) of the high-level policy includes Cartesian Space commands of the left and right hands $a_{\text{lh}}^3, a_{\text{rh}}^3 \in \mathcal{R}^3$ and torso yaw joint position commands $a_{\text{ty}}^3 \in \mathcal{R}^1$. These actions $a^3$ were executed by the mid-level stability controller.

The states $s^3 \in \mathcal{S}^3 \subseteq \mathcal{R}^{14}$ consists of the joint positions of the arms ($s_{joints}{}^3 \in \mathcal{R}^8$), Cartesian positions of the box ($s_{box}{}^3 \in \mathcal{R}^3$), and three Boolean values whether the box is in contact with the hands, table, and whether the button is pushed.

A reward is given for three cases: (1) box in hands ($r_{bih}$), (2) box on conveyor belt($r_{boc}$), and (3) button pushed ($r_{pb}$) while the box is on the second conveyor belt. The resulting reward function with weights $w_i$ (see Table S2 bottom) are:

$$r = w_1 r_{\text{bih}} + w_2 r_{\text{boc}} + w_3 r_{\text{bp}}.$$

**Mid-level Stability Control**

The mid-level stability control level consisted of two components: the manipulation policy was realised as Model-Predictive Control (MPC) scheme for the arms, and a locomotion policy was learned through Deep Reinforcement Learning for the legs.

***Manipulation Policy***

As input into the policy, the manipulation policy received Cartesian target positions for the hands $a^3 = [a_{\text{lh}}, a_{\text{rh}}]$ from the high-level policy, current Cartesian position of the hands $s^2 = [p_{\text{lh}}, p_{\text{rh}}] \in \mathcal{S}^2 \subseteq$ $\mathcal{R}^6$, and current, measured joint angles of the arms $o \subseteq \mathcal{R}^8$. The output $a^2 = q_{\text{arms}}^d \in \mathcal{A}^2 \subseteq \mathcal{R}^8$ of the manipulation policy was target joint positions $q_{\text{arms}}^d$ of the arms to the Low-level Joint Controller.

The manipulation policy consists of two parts (see flow diagram in Fig. S2): Model-Predictive Control (MPC) that generated a stable, optimal trajectory in Cartesian space and Inverse Kinematics (IK)[79] that transformed desired actions from the Cartesian space to the joint space.

To provide the smoothest possible motions for the hands, we formulated the optimal control problem as the minimum-jerk optimisation, while satisfying dynamics constraints on the hands. The optimal trajectory was then implemented in an MPC fashion. The MPC control applied the first control input of the optimal input trajectory and then re-optimised based on the new state at the next control loop[80]. In this way, MPC successively solved an optimal control problem over a prediction horizon $N$ and achieved feedback control, while ensuring optimality.

For the hand position $p$, an objective function $J$ was designed to minimise jerk $\dddot{p}$ (the input $u$ of the system) with final time $t_f$:

$$J = \frac{1}{2} \int_0^{t_f} \left( \frac{\mathrm{d}^3 p(t)}{\mathrm{d}t^3} \right)^2 \mathrm{d}t = \frac{1}{2} \int_0^{t_f} u(t)^2 \mathrm{d}t.$$

The Minimum Jerk MPC (MJMPC) solved the following constrained optimisation problem at every time step at a frequency of $25Hz$:

$$\begin{aligned}
\min_{u(t)} \quad & \frac{1}{2} \int_0^{t_f} u(t)^2 \mathrm{d}t \\
\text{subject to} \quad & \frac{\mathrm{d}^3 p(t)}{\mathrm{d}t^3} = u \\
& [p(0), \dot{p}(0), \ddot{p}(0)] = [p_0, \dot{p}_0, \ddot{p}_0] \\
& [p(t_f), \dot{p}(t_f), \ddot{p}(t_f)] = [p_f, \dot{p}_f, \ddot{p}_f] \\
[p_{\min}, \dot{p}_{\min}, \ddot{p}_{\min}] \leq & [p, \dot{p}, \ddot{p}] \leq [p_{\max}, \dot{p}_{\max}, \ddot{p}_{\max}],
\end{aligned}$$

with initial condition $[p_0, \dot{p}_0, \ddot{p}_0]$, and terminal condition $[p_f, \dot{p}_f, \ddot{p}_f]$.

The resultant Cartesian trajectory $p^d$, i.e., the trajectory that leads from the initial hand position $p_0$ to the final hand position $p_{tf}$, from MJMPC was transformed into joint position commands $q_{\text{arms}}^d$ through Inverse Kinematics (IK). More formally, IK described a transformation $T: \mathcal{C} \to \mathcal{Q}$ from Cartesian space $\mathcal{C}$ to joint space $\mathcal{Q}$. The joint position commands $q^d$ were then tracked by the low-level joint position controller as described in Section "Low-level Joint Control". The IK ensures feasible joint configuration on the robot even if the high-level decision policy or the MPC trajectory yield infeasible setpoints.

***Locomotion Policy***

The locomotion policy $\pi(s; \theta)$ coordinated the 12 Degree of Freedom (DoF) leg joints and was instantiated as a Deep Neural Network (network parameters $\theta$) that received robot states $s$ as inputs and outputs 12 target joint positions $q^d_{\text{legs}}$ for the legs. It was trained through Soft-Actor Critic (SAC) [81] – an off-policy Deep Reinforcement Learning (DRL) algorithm.

SAC optimised a maximum entropy objective $J_{\text{SAC}(\pi)}$:

$$J_{\text{SAC}}(\pi) = \sum_{t=0}^{T} \mathbb{E}\left[r(s_t, a_t) + \alpha\mathcal{H}\big(\pi(\cdot \,|s_t)\big)\right],$$

with reward $r$, state $s_t$ and action $a_t$ at time $t$, temperature parameter $\alpha$, and policy entropy $\mathcal{H}(\pi)$. The parameters $\theta$ for policy $\pi_\theta$ were obtained by minimising $J_\pi(\theta)$:

$$J_\pi(\theta) = \mathbb{E}\big[\log\pi_\theta(a_t|s_t) - Q_\phi(s_t, a_t)\big].$$

The action-value function $Q_\phi(s_t, a_t)$ was obtained by minimising the Bellman residual $J_Q(\phi)$:

$$J_Q(\phi) = \mathbb{E}\left[1/2\left(Q_\phi(s_t, a_t) - \hat{Q}(s_t, a_t)\right)^2\right],$$

with Bellman equation $\hat{Q}(s_t, a_t) = r(s_t, a_t) + \gamma\mathbb{E}\big[V_\psi(s_{t+1})\big]$ and discount factor $\gamma$. Estimation of the value function $V_\psi$ was obtained through minimising $J_V(\psi)$:

$$J_V(\psi) = \mathbb{E}[1/2\left(V_\psi(s_t) - \mathbb{E}\big[Q_\theta(s_t, a_t) - \log\pi_\phi(a_t|s_t)\big]\right)^2].$$

The training procedures, including the design of reward, action space, and state-space, are as in [77]. The actions $a^2 \in \mathcal{A}^2 \subseteq \mathcal{R}^{12}$ were the joint positions $q_{\text{legs}}$ of the 12 Degree of Freedoms (DoF) of the legs (for each leg 3 DoF hip, 1 DoF knee, 2 DoF ankle). The target joint positions $q_{\text{legs}}^d$ were tracked by the low-level joint controller (see Section "Low-level Joint Control").

The state $s^2 \in \mathcal{S}^2 \subseteq \mathcal{R}^{27}$ consisted of the target pelvis position $a_{\text{pelvis}}^3$ (the walking destination), proprioceptive information of the robot including pelvis orientation, linear and angular velocity of the pelvis, force of both feet, joint positions of the legs, and the gait phase. The gait phase indicates the phase of the periodic gait at any point in time, which is implemented as a two-dimensional vector on the unit-circle to describe the phase of periodic trotting. For more details regarding the gait phase state, please refer to[77], where the gait phase is used to enable the imitation learning of periodic locomotion.

The reward comprised of an imitation term and a task term:

$$r = w_i r_{\text{imitation}} + w_t r_{\text{task}},$$

where weights $w_i, w_t$ and reward terms $r_{\text{imitation}}, r_{\text{task}}$ for imitation and task respectively. The imitation term encourages human-like motions by rewarding motions that are close to a reference motion capture trajectory. The task reward term rewards motions that contribute towards achieving the task, i.e., walking towards a goal while maintaining balance.

To encourage a state $x$ to be close to a desired target value $\hat{x}$, the corresponding reward component was designed as Radial Basis Function (RBF) kernel $K(\hat{x}, x, \alpha)$:

$$K(\hat{x}, x, \alpha) = e^{-\alpha(\hat{x}-x)^2},$$

with hyperparameter $\alpha$ controlling the width of the kernel.

The aim of $r_{\text{imitation}}$ was to imitate the joint position, feet pose, and contact pattern of a reference motion capture trajectory as closely as possible. This is achieved by the reward function $r_{imitation}$:

$$r_{imitation} = w_{joint\_position} r_{joint\_position} + w_{pose} r_{pose} + w_{contact} r_{contact}.$$

The reward components $r_{joint\_position}, r_{pose}$ use the RBF kernel to encourage the policy learning motions that are close to the reference joint positions and feet poses respectively. The contact reward $r_{contact}$ is a binary reward that is one if the foot in the reference motion was in contact with the ground and zero otherwise. The weights used for the reward components can be found in Table S3. The target references for joint position, feet pose, and feet contact come from the motion capture study[82].

The reward term $r_{task}$ rewarded upright posture, short distances to the goal position, and regularised the joint velocity and torque:

$$r_{task} = w_{pose}r_{pose} + w_{goal}r_{goal} + w_{vel}r_{vel} + w_{torque}r_{torque},$$

With the values of the weights $w_{pose}, w_{goal}, w_{vel}, w_{torque}$ as in Table S3, and reward components $r_{pose}, r_{goal}, r_{joint\_vel}, r_{torque}$ that respectively reward the torso pose to be upright, the distance vector between pelvis and goal to be as small as possible, and the joint velocity and joint torque to be as small as possible. The RBF kernel is used for all reward components in $r_{task}$.

**Low-level Joint Control**

The low-level joint control tracked the target joint positions $q^d = [q^d_{arms}, q^d_{legs}]$ provided by the mid-level stability controller (see flow diagram in Fig. S3). It receives joint positions $q \in \mathcal{R}^{20}$, joint velocities $\dot{q} \in \mathcal{R}^{20}$ and target joint position targets $q^d = a^2 \in \mathcal{R}^{20}$ as input and outputs motor current $a^1 = I \in \mathcal{A}^1 \subseteq \mathcal{R}^{20}$.

It was implemented as *Joint Impedance Controller* that regulated around the set point to achieve accurate tracking of the desired joint motions $q^d$.

The joint impedance control calculated the desired joint torque $\tau^d$ using position $q$ and its derivative $\dot{q}$, with the stiffness $K_{P_i}$ and damping $K_{D_i}$ gains:

$$\tau^d = K_{P_1}(q^d - q) - K_{D_1}\dot{q}.$$

At the actuator level, the motor driver implemented an internal current control to track the desired joint torque $\tau^d$ using a proportional-derivative law, where the desired motor current $I$ was computed as:

$$I = K_{P_2}(\tau^d - \tau) - K_{D_2}\dot{\tau}.$$

# DATA AVAILABILITY

The data analysed in this work was generated using the code provided in our open-source repository, where source data is also provided. Further information can be found in our repository (see the Code Availability section) and in the repository
https://gitfront.io/r/yunaik/363391223108e06f2df66e2151bacf568bf3b979/nmi-fig-data/

# CODE AVAILABILITY

The code used in this work is available on
https://gitfront.io/r/yunaik/fdbaf62d7d63283c66a1fe66713775f8fb37d7fd/nmi-val/

# AUTHOR CONTRIBUTIONS

K.Y. and Z.L. conceptualised the robot control architecture. K.Y., N.S., and K.F. designed and formulated the hierarchical generative model. K.Y. implemented the model and performed the robotic experiments. K.Y. and N.S. wrote the manuscript. All authors contributed to and edited the manuscript.

# COMPETING INTERESTS

# Hierarchical generative modelling for autonomous robots

Kai Yuan[1,4], Noor Sajid[2,4], Karl Friston[2] and Zhibin Li[3*]

[1] Embodied AI Lab, Intel Labs, Germany
[2] Wellcome Centre for Human Neuroimaging, Queen Square Institute of Neurology, University College London, UK
[3] Computer Science & Robotics Institute, University College London, UK
[4] These authors contributed equally.

* Corresponding author: *alex.li@ucl.ac.uk*

# ROBOT PLATFORM

The motions for autonomous task completion are implemented on NASA's humanoid Valkyrie[13]. Valkyrie was designed to operate in extra-terrestrial planetary space missions such as unmanned pre-deployment on Mars – it is 1.87m tall, consists of 44 Degrees of Freedom, and weighs 129kg with ranges of motions similar to humans. The 25 series-elastic actuators in arms, torso, and legs enable human-like locomotion and manipulation; all the joint limits are detailed in Table S1. Valkyrie can sense the environment through proprioceptive and exteroceptive sensors, including a multitude of gyroscopes, accelerometers, load cells, pressure sensors, sonar, LIDAR, depth cameras, and stereo sensors. In this work, the proprioceptive information was measured by simulated onboard sensors. Signals originating from exteroceptive sensors, such as the relative positions between the robot and other objects in the environment are directly queried from the simulation.

**Table S1:** Mechanical specifications of Valkyrie.

| | Lower joint position [rad] | Upper joint position [rad] | Joint velocity [rad/s] | Joint torque [Nm] |
|---|---|---|---|---|
| Shoulder roll | -1.3 | 1.5 | 5.9 | 190 |
| Shoulder pitch | -2.9 | 2.0 | 5.9 | 190 |
| Shoulder yaw | -3.1 | 2.2 | 11.6 | 65 |
| Elbow pitch | -2.2 | 0.1 | 11.5 | 65 |
| Torso roll | -0.2 | 0.3 | 9.0 | 150 |
| Torso pitch | -0.1 | 0.7 | 9.0 | 150 |
| Torso yaw | -1.3 | 1.2 | 5.9 | 190 |
| Hip Roll | -0.6 | 0.5 | 7.0 | 350 |
| Hip pitch | -2.4 | 1.6 | 6.1 | 350 |
| Hip yaw | -0.4 | 1.1 | 5.9 | 190 |
| Knee pitch | -0.1 | 2.1 | 6.1 | 350 |
| Ankle Pitch | -0.9 | 0.7 | 11 | 205 |
| Ankle Roll | -0.4 | 0.4 | 11 | 205 |

**Table S2:** Weights for high-level reward. Top: Task 1, Bottom: Task 3

| | $w_1$ | $w_2$ | $w_3$ | $w_4$ | $w_5$ | $w_6$ |
|---|---|---|---|---|---|---|
| Value | 1 | 0.1 | 1 | 2 | 2 | 5 |

| | $w_1$ | $w_2$ | $w_3$ |
|---|---|---|---|
| Value | 1 | 5 | 10 |

**Table S3:** Weights for mid-level reward

| | $w_i$ | $w_t$ | $w_{pose}$ | $w_{goal}$ | $w_{vel}$ | $w_{torque}$ | $w_{joint\ position}$ | $w_{pose}$ | $w_{contact}$ |
|---|---|---|---|---|---|---|---|---|---|
| Value | 0.5 | 0.5 | 0.15 | 0.25 | 0.5 | 0.1 | 0.5 | 0.4 | 0.1 |

**Table S4:** Summary of Implicit Hierarchical Generative Model for a robotics system.

| | Joint Probability Distribution | Algorithm | Input States | Outcomes | Function |
|---|---|---|---|---|---|
| **High-Level** | $P(s_{t_3}^3 \mid s_{t_3-1}^3, a_{t_3-1}^3, o_{t-1})$ | Double Q-Learning | 1. Relative position to table, 2. Relative position to box | 1. Pelvis target, 2. Left and right-hand target | Provide targets for locomotion and manipulation |
| **Mid-Level** | $P(s_{t_2}^2 \mid s_{t_2-1}^2, a_{t_2}^3, a_{t_2-1}^2, o_{t_1-1})$ | Soft Actor-Critic, Minimum-Jerk Model-Predictive Control | 1. Pelvis targets, 2. Left and right-hand target, 3. Gait phase proprioceptive robot states | 1. Joint positions for legs and arms | Coordinate arm and leg joints to perform locomotion and manipulation |
| **Low-Level** | $P(s_{t_1}^1 \mid s_{t_1-1}^1, a_{t_2}^2, a_{t_1-1}^1, o_{t_1-1})$ | Joint Impedance Control | 1. Joint position targets, 2. Proprioceptive robot states | 1. Motor current for robot joints | Actuation of robot joints |

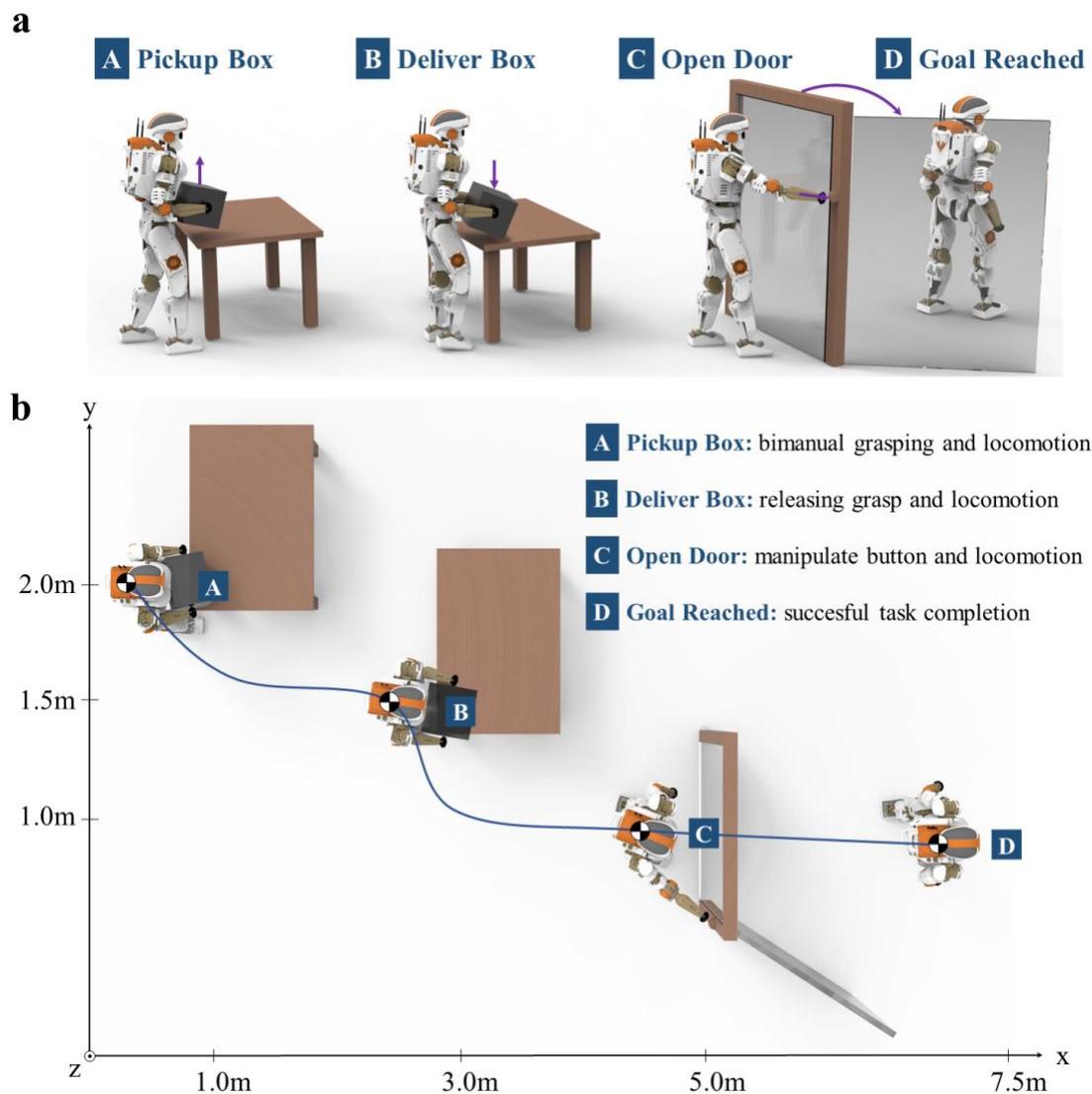

**Figure S1:** Visual presentation of the robot task: pick up a box, deliver it to the second table, approach the door, press the button to open the door, and enter the destination once the door opens fully. (a) shows the side view of the task; (b) shows the top-down view with x-y coordinates in Cartesian space.

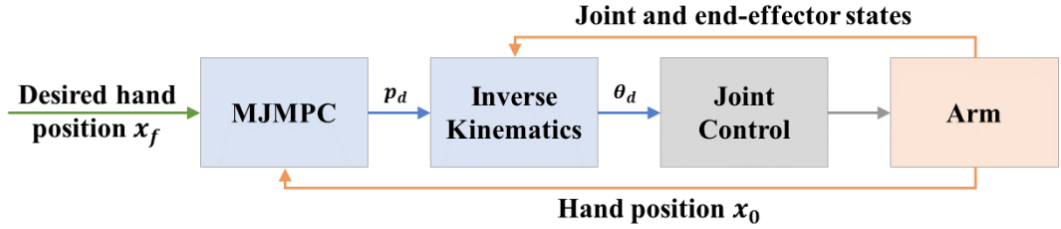

**Figure S2:** Control diagram of the manipulation policy. The Minimum Jerk Model-Predictive Control scheme provides a Cartesian space trajectory which is transformed into the joint space by Inverse Kinematics.

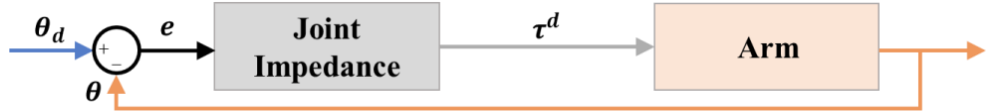

**Figure S3:** Control structure for low-level joint control: Joint Impedance Control (feedback).

---

**Algorithm 1** Pseudocode of Implicit Hierarchical Generative Model

---

1: **while** Task ongoing **do**
2:     Get robot states $s_t^3$
3:     Get pelvis and hand targets via (x)
4:     **for** mid-level loop duration=1,...,50 **do**
5:         Get robot states $s_t^2$
6:         Get leg joint position that track pelvis target using locomotion policy (x)
7:         Get arm joint position that track hand targets using manipulation policy (x)
8:         **for** low-level loop duration=1,...,20 **do**
9:             Get robot states $s_t^1$
10:            Get motor joint currents that track leg and arm joint positions using joint impedance control (x)

---

**Figure S4:** Pseudocode of Implicit Hierarchical Generative Model.

**a1** Right hand motion in Cartesian space

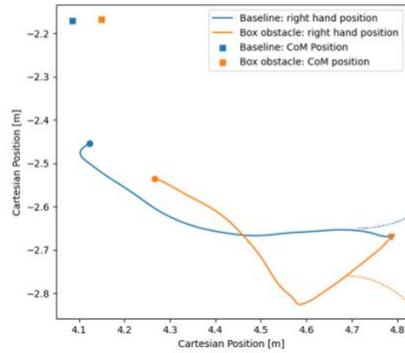

**a2** Baseline reach motion

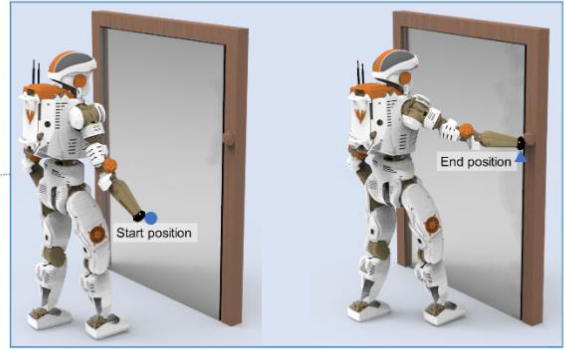

**b** Right arm motion in Joint space

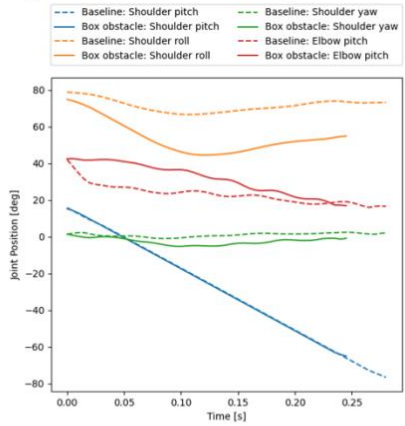

**a3** Box obstacle: reach motion

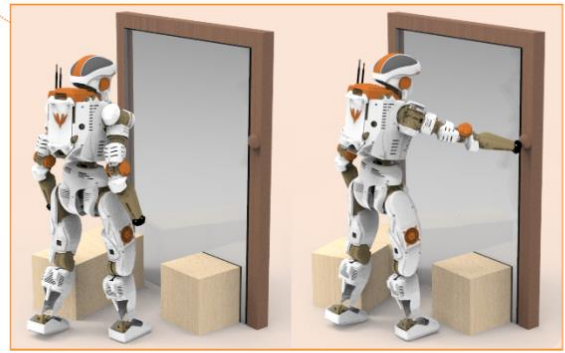

**Figure S5:** Adaptive arm movements and manipulation during the door-opening task. Panel **a** shows the right arm's motion in cartesian space for the baseline (no obstacle) and the box obstacle scenario. When there is an obstacle in front of the door, the arm motion must adapt, as the robot's upper body position deviates from its baseline position. The multi-joint coordination aspect can be seen in Panel **b**. The adapts to the change of scenario, coordinating the multiple arm joints, such that the end position of the arm is pressing the button to open the door, while being obstructed by box obstacles.

# COMPARISON BETWEEN HIERARCHICAL AND FLAT ARCHITECTURES

To demonstrate that Hierarchical Generative Models (HGM) are more effective for modelling complex and sequential data than flat generative models, we design a flat architecture and compare both models on Task 1 " Box Delivery and Opening Door". The flat architecture has the same amount of weights as the HGM without the hierarchical component. The state space of the policy is the combined state space of high-level, mid-level, and low-level policy of HGM. The action space is direct torque control on all 25 actuators (arms, torso, and legs). The flat architecture is thus end-to-end learning both the high-level decision making and the low-level motor control by receiving both exteroceptive and proprioceptive information, while also receiving task-relevant states.

While HGMs can learn to complete tasks, such as box delivery, opening a door, and penalty kicks, flat generative models fail to learn both the decision making and the motor coordination aspect. The learning curves show how the flat architecture (Fig. S6a top) is unable to learn while HGM (Fig. S6a bottom) converges after 1e6 steps. A snapshot showing how the flat architecture fails to learn to balance and locomote can be seen in Fig. S6b. This is due to several key differences in the way these models address the data generation.

Firstly, HGM structures the data generation process into different levels, allowing for a more abstract representation of the data and a greater degree of control over the generation process. Such a hierarchical structure leads to a better understanding of the underlying structure of the data, resulting in more efficient and effective data generation. In contrast, flat generative models lack a hierarchical structure, leading to a cluttered and less manageable representation of the data.

Secondly, another advantage of HGM is that it employs a higher-level policy that decides which lower-level policy to execute. This allows for a more modular and scalable approach to modelling complex sequential data. In contrast, flat generative models rely on a single policy for all decisions.

Finally, HGM has the ability to leverage prior knowledge and skills learned at lower levels of the hierarchy to tackle new challenges at higher levels. This process of building upon prior knowledge and skills allows HGM algorithms to model more quickly and effectively than flat generative model algorithms. In contrast, flat generative models start from scratch for each new data set, leading to longer training times and a higher risk of overfitting to specific instances of the data. The ability to reuse prior knowledge and skills in HGM leads to a more sample-efficient and flexible data generation process, enabling better adaptation to new and changing data sets.

**a**

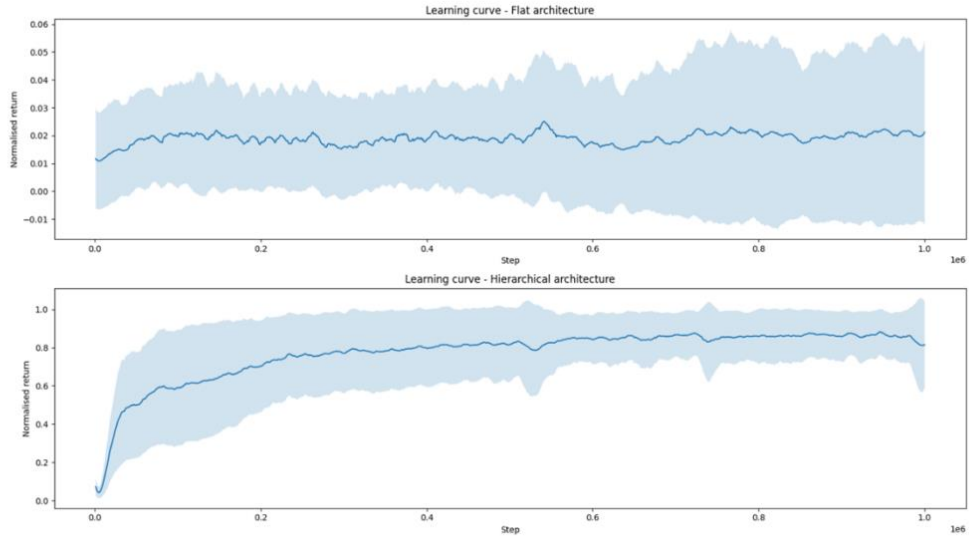

**b**

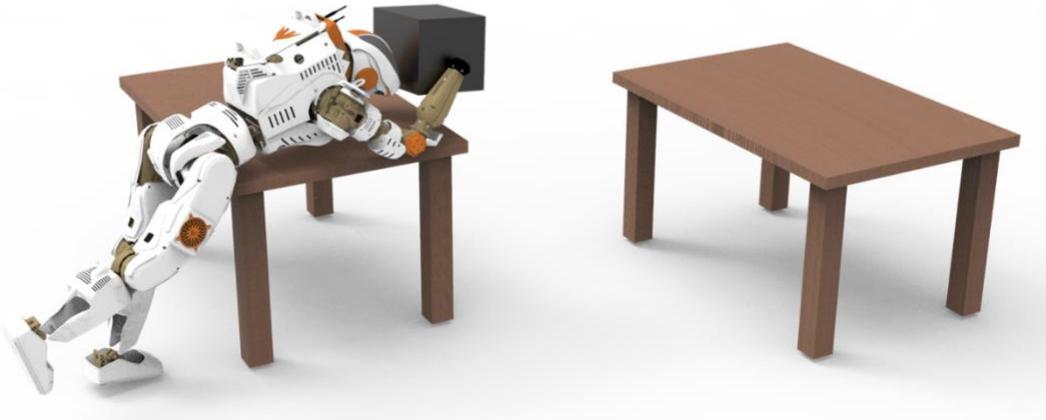

**Figure S6:** Panel a shows the learning curve of flat (top) and hierarchical (bottom) structure. Panel b illustrates an unsuccessful robot policy due to the inability to learn how to maintain balance.